\documentclass[10pt,twocolumn]{article}
\usepackage[margin=0.75in,columnsep=0.25in]{geometry}
\usepackage{amsmath,amsfonts,amssymb}
\usepackage{graphicx}
\usepackage{booktabs}
\usepackage{mdframed}
\usepackage{microtype}      
\usepackage[numbers]{natbib}
\usepackage{listings}
\usepackage{fancyvrb}
\usepackage[hyphens]{url}
\usepackage{hyperref}
\usepackage[hyphenbreaks]{breakurl}

\newcommand{\class}[2]{\mbox{\textbf{\lstinline{#2}}.\lstinline|#1|}}

\title{FedRAG: A Framework for Fine-Tuning Retrieval-Augmented Generation Systems}
\author{
Val Andrei Fajardo$^1$ \and
David B. Emerson$^1$ \and  
Amandeep Singh$^1$ \and
Veronica Chatrath$^1$ \and
Marcelo Lotif$^1$ \and
Ravi Theja Desetty$^2$ \and
Chi Ho Cheung$^2$ \and
Izuki Matsuba$^2$ \\
\\
$^1$Vector Institute, Toronto ON M5G 0C6, Canada \\
$^2$Independent Researcher, Toronto, Canada
}
\date{}

\begin{document}
\maketitle

\begin{abstract}

Retrieval-augmented generation (RAG) systems have been shown to be
effective in addressing many of the drawbacks of relying solely on the parametric memory
of large language models.  Recent work has
demonstrated that RAG systems can be improved via fine-tuning of their retriever and generator models.  In this work, we introduce FedRAG, a framework
for fine-tuning RAG systems across centralized and federated architectures.
FedRAG supports state-of-the-art fine-tuning methods, offering a simple and
intuitive interface and a seamless conversion from centralized to federated
training tasks. FedRAG is also deeply integrated with the modern RAG ecosystem,
filling a critical gap in available tools.

\end{abstract}

\section{Introduction}

Large Language Models (LLMs) have demonstrated a remarkable capacity to perform
a diverse set of tasks, despite standard pre-training only incorporating a
next-token prediction objective~\cite{Brown1, Grattafiori1,
chhun-etal-2022-human, pmlr-v239-ren23a, paul-etal-2024-refiner}. Prompt
engineering techniques, such as in-context learning, chain-of-thought (CoT), and
self-consistency have been shown to further improve the performance of
pre-trained LLMs in many settings~\cite{Brown1, cot1, WangSelfConsistency}.
However, there are tasks, particularly knowledge-intensive ones, where relying
solely on information encoded in an LLM's parameters, often referred to as
parametric memory, may lead to factual inaccuracies in model responses,
colloquially named hallucinations~\cite{Huang1}. Hallucinations issues can become more prevalent when
models attempt to address queries beyond their knowledge
cutoffs~\cite{dige2024evaluating,
ovadia2024finetuningretrievalcomparingknowledge}.

Retrieval-augmented generation (RAG) is a popular methodology that aims to address this
specific drawback~\cite{LewisRAG1, ram-etal-2023-context}. Specifically,
RAG systems present relevant non-parametric knowledge drawn from external
systems alongside queries in the form of additional input to an LLM. Perhaps the most widely used form of this involves computing an embedding
of right-sized chunks of the non-parametric knowledge and storing these in a vector
store like Qdrant, Chroma or Pinecone for future search and retrieval~\cite{Fan1}.
More elaborate designs of RAG systems have also been developed, such as
those utilizing knowledge graphs~\cite{peng2024graphretrievalaugmentedgenerationsurvey}.

Recent studies have shown that fine-tuning RAG systems can lead to even greater
performance improvements~\cite{lin2023ra, zhang2024raft, chen2025research}.
That is, through fine-tuning, the overall RAG system, comprised of a generator,
retriever and knowledge store, may be adapted to work more cohesively as a
single unit in performing tasks.

The RAG ecosystem of today is quite vibrant and includes a wide range of options
for LLMs, retrieval models, and re-rankers, among other
components~\cite{karpukhin-etal-2020-dense, ma-etal-2023-query, reranking1,
wang2023knowledgegraphpromptingmultidocument, zhuang-etal-2023-open} being
offered by organizations operating under both open- as well as closed-source
business models. There are also more than a few storage, observability, and
evaluation solutions that developers can choose from in order to build an end-to-end
RAG production. Finally, popular RAG frameworks such as
LlamaIndex~\cite{Liu_LlamaIndex_2022} and LangChain~\cite{Chase_LangChain_2022}
offer users the ability to rapidly assemble and experiment with diverse RAG
system configurations, ultimately aiding in the discovery of optimal designs.
Yet, to the best of our knowledge, there are few, if any, frameworks that help
simplify RAG fine-tuning, while remaining well integrated with other available
tools and resources in the ecosystem.

The work presented here aims to directly fill this gap. Specifically, we
introduce FedRAG, a framework for fine-tuning RAG systems across both
centralized and federated architectures.\footnote{Library Code: \url{https://github.com/VectorInstitute/fed-rag}} Decentralized designs for LLM training
and deployment are becoming increasingly important, as evidenced by popular
initiatives like Anthropic's Model Context Protocol (MCP)~\cite{anthropic2024mcp} and Google's Agent2Agent Protocol~\cite{google2025a2a}.
Moreover, in settings where data privacy prevents centralizing datasets, decentralized training techniques like federated learning (FL) become an indispensable tool for improving
RAG systems.

\section{Related Work}

\subsection{Fine-tuning RAG Systems}

As discussed above, RAG systems are comprised of a number of components, some of
which are driven by trainable models. This work specifically focuses on two main
components: the generator, which is responsible for text generation; and the
retrieval model, which maps queries or prompts into a form, commonly a high-dimensional
embedding vector, used to retrieve related context from a knowledge store.

Several studies have focused on generator training via instruction fine-tuning,
for which the instruction examples include context retrieved by the retriever
from the knowledge store.~\citet{lin2023ra} refer to this approach as
Retrieval-Augmented Language Model Training (RALT). A similar generator
fine-tuning approach called Retrieval-Augmented Fine-Tuning (RAFT) was
utilized in~\cite{zhang2024raft}. RAFT differs from RALT in that the instruction
examples also include LLM generated CoT passages with reasoning traces linking the response to the query. Finally, in line with recent trends in
reasoning LLMs~\cite{kumar2025llmposttrainingdeepdive},~\citet{chen2025research}
introduce ReSearch, which follows a reinforcement learning approach similar to
that used in
DeepSeek-R1~\cite{deepseekai2025deepseekr1incentivizingreasoningcapability}.
With ReSearch, the LLM learns to generate long CoT passages that incorporate
search and retrieval from a knowledge store. In so doing, the generator is
adapted to cycle between reasoning and retrieval, potentially multiple times.

Fewer studies exist considering retriever fine-tuning. In the same work that
produced RALT, the authors also introduce Language
Model Supervised Retriever Training (LSR). In LSR, the retrieval scores of
retrieved text chunks as well as corresponding target sequence probabilities
produced by the generator model, conditioned on the context of each retrieved chunk, form two distributions whose distance, measured by the Kullback-Leibler
divergence, is minimized.

\subsection{Federated Learning for LLM Applications}

Recently, Flower Labs developed the first federally pre-trained LLM called FlowerLLM.\ Further, efforts supporting federated LLM
tuning have been undertaken~\cite{sani2024photonfederatedllmpretraining}. In the
context of RAG systems, however, work has primarily focused on decentralized
inference rather than federated fine-tuning or training. A notable example
published by Flower Labs demonstrates RAG inference in a federated
setting.\footnote{\url{https://flower.ai/docs/examples/fedrag.html}} Similarly,
LlamaIndex has also developed a library extension to their framework called
\texttt{llama-index-networks}~\cite{Fajardo_LlamaIndexNetworks_2024} for
decentralized RAG inference. To the best of our knowledge, no existing tools
provide a simple interface for converting centralized RAG fine-tuning to
federated tasks.

\section{Philosophy and Design Principles}

In this section, we describe the core philosophy and design principles that
guide the development of FedRAG\@. These principles address the challenges
identified in the previous section and inform implementation decisions.

\subsection{Philosophy}

We endeavour to build FedRAG for researchers and practitioners alike in a manner
that makes applying state-of-the-art fine-tuning techniques to RAG systems
simple, yet highly effective, irrespective of whether the system is centralized
or federated. Moreover, we seek to make researching new methods for RAG
fine-tuning more efficient and scientifically rigorous by promoting
reproducibility. This is achieved through designing flexible components and
appropriate tools that allow users to easily replicate and disseminate their RAG
systems, as well as extend the library with custom trainers, losses, benchmarks,
and more.

\subsection{Design Principles}\label{sec:design-principles}

\paragraph{Advanced RAG Fine-Tuning:} \emph{Comprehensive support for
state-of-the-art RAG fine-tuning methods that can be federated with ease.}

This principle is central to advancing the state of knowledge in RAG methods. By
implementing and supporting frontier techniques in RAG fine-tuning, while
simultaneously making federation straightforward and accessible, researchers are
enabled to develop and evaluate novel approaches in a systematic and
reproducible fashion. At the same time, such methods transfer smoothly to
decentralized systems.

\paragraph{Work With Your Tools:} \emph{Seamless integration with popular
frameworks including HuggingFace, Unsloth, and LlamaIndex, to become deeply
embedded in the RAG ecosystem and beyond.}

By designing FedRAG as a deeply embedded framework within the existing RAG
ecosystem, barriers to adoption are significantly reduced for both practitioners
and researchers.  Integrations into popular frameworks and libraries, such as those
mentioned above, allow users to leverage familiar tools and workflows while
gaining access to advanced fine-tuning capabilities. Finally, extensive ecosystem
compatibility facilitates discoverability and further adoption of new methods
and results, thus maximizing the impact and reach of research advancements.

\paragraph{Lightweight Abstractions:} \emph{Clean, intuitive abstractions that
simplify RAG fine-tuning while maintaining full flexibility and control.}

We seek to provide developers with an intuitive interface and abstractions that
are easy to work with, customize, and extend. Lowering the learning curve to use
FedRAG, while simultaneously increasing its utility and effectiveness, is
essential to providing a pleasant development experience. This approach enables
practitioners and researchers to focus their efforts entirely on the challenges
of designing and experimenting with methods for improving RAG systems rather
than wrestling with complex implementation details.

\section{FedRAG Framework Overview}

This section provides a more detailed overview of the FedRAG library.

\subsection{Library Organization}

FedRAG incorporates a modular design, consisting of several modules with clear and
intuitive separation of concerns. Table \ref{tab:modules} presents non-exhaustive overview of the key modules
and their responsibilities.

\begin{table}[ht]
\centering
  \footnotesize
  \begin{tabular}{@{}ll@{}}
  \toprule
  Module & Description \\
  \midrule
    \textbf{\lstinline{core}} & Core types i.e., \lstinline|RAGSystem| \\
    \textbf{\lstinline{evals}} & Evaluation metrics and benchmarks \\
    \textbf{\lstinline{fl\_tasks}} & Federated learning task definitions \\
    \textbf{\lstinline{generators}} & Generator types \\
    \textbf{\lstinline{knowledge\_stores}} & Data storage \\
    \textbf{\lstinline{retrievers}} & Retriever types \\
    \textbf{\lstinline{trainers}} & Trainer types \\
    \bottomrule
  \end{tabular}
  \caption{Key modules in FedRAG and their responsibilities.}
  \label{tab:modules} 
\end{table}

\subsection{Standard Usage Patterns}

\paragraph{Building a RAG System:} We first introduce the main classes that
FedRAG offers and with which users will often work. We begin with the
\class{RAGSystem}{core} class, which is comprised of three parts, namely:
\class{KnowledgeStore}{knowledge\_stores}, \class{Retriever}{retrievers}, and
\class{Generator}{generators}. Figure~\ref{lst:ragsystem} provides a code
snippet on how to assemble a \lstinline|RAGSystem| with FedRAG.

\begin{figure}[ht!]
\begin{mdframed}[linewidth=1pt]
\begin{Verbatim}[fontsize=\small]
# Flat imports are supported
from fed_rag import (
  RAGSystem,
  RAGConfig,
  HFSentenceTransformerRetriever,
  UnslothFastModelGenerator,
  QdrantKnowledgeStore
)

knowledge_store = QdrantKnowledgeStore()
generator = UnslothFastModelGenerator(
  "unsloth/gemma-3-4b-it",
)
retriever = HFSentenceTransformerRetriever(
  "nthakur/dragon-plus-query-encoder",
  "nthakur/dragon-plus-context-encoder",
)

# Assemble rag_system
rag_system = RAGSystem(
  knowledge_store=knowledge_store,
  generator=generator,
  retriever=retriever,
  rag_config=RAGConfig(top_k=2)
)

# Executes the typical RAG pipeline
response = rag_system.query("What are tulips?")
\end{Verbatim}
\end{mdframed}
\vspace{-3ex}
\caption{Creating a \lstinline|RAGSystem| with FedRAG. Not depicted here,
but the \lstinline|retriever| is also used to populate embedded reference chunks into the
\lstinline|knowledge_store|, prior to querying.}\label{lst:ragsystem}
\end{figure}

\paragraph{RAG Fine-Tuning:} After the essential RAG components are constructed, the system
can be trained using the fine-tuning abstractions offered by the library. FedRAG
provides various \class{Trainer}{trainers} classes distinguished by their
methodology and which model, generator or retriever, they target. A typical
pattern for performing retriever or generator training is provided in
Figure~\ref{lst:fine-tuning-ragsystem}. There, the \lstinline|manager| is an orchestrator object that bears the
responsibility of preparing the target model for training and ensuring the other
model is frozen. Note that both generator and retriever trainer objects also
expose a \mbox{\lstinline|train()|} method that can be called without using
\lstinline|manager| providing an interface similar to that of
HuggingFace.

\begin{figure}[ht!]
\begin{mdframed}[linewidth=1pt]
\begin{Verbatim}[fontsize=\small]
... # Keep code from Figure 1
from fed_rag.trainers import (
  HuggingFaceTrainerForRALT
  HuggingFaceTrainerForLSR
)
from fed_rag.trainer_managers import (
  HuggingFaceRAGTrainerManager
)
from datasets import Dataset

# Train datasets are examples of (query, 
# response) pairs
train_dataset = Dataset.from_dict(
  {
    "query": [...],
    "response": [...]
  }
)
generator_trainer = HuggingFaceTrainerForRALT(
    rag_system=rag_system,
    train_dataset=train_dataset,
)
retriever_trainer = HuggingFaceTrainerForLSR(
    rag_system=rag_system,
    train_dataset=train_dataset,
)
manager = HuggingFaceRAGTrainerManager(
    mode="retriever",  # can be generator
    retriever_trainer=retriever_trainer,
    generator_trainer=generator_trainer,
)

# Train
train_result = manager.train()
\end{Verbatim}
\end{mdframed}
\vspace{-3ex}
\caption{Fine-tuning a \lstinline|RAGSystem| with FedRAG.}\label{lst:fine-tuning-ragsystem}
\end{figure}

\paragraph{Federated Fine-Tuning:} With the centralized fine-tuning pattern
established, we show the simple process for converting the previous task to a
federated one. This is done by extracting an \class{FL_task}{fl_tasks} object
from the \lstinline|manager|. This is demonstrated in
Figure~\ref{lst:fed-fine-tuning-ragsystem}. Each client has its own fine-tuning dataset and contribute to the tuning process in a decentralized manner and updates are combined with federated averaging~\cite{McMahan1}.

\begin{figure}[ht!]
\begin{mdframed}[linewidth=1pt]
\begin{Verbatim}[fontsize=\small]
... # Keep code from Figures 1 and 2
import flwr as fl  # The FL backend
# fl_task
fl_task = manager.get_federated_task()
# Build fl server and client
server = fl_task.server(
  model=retriever_trainer.model
)
client = fl_task.client(
  model=retriever_trainer.model,
  train_dataset=train_dataset,
)
# Spin up client and server using flwr
fl.start_server(server)
fl.start_client(client)
\end{Verbatim}
\end{mdframed}
\vspace{-3ex}
\caption{Federated fine-tuning of a \lstinline|RAGSystem| with FedRAG.}\label{lst:fed-fine-tuning-ragsystem}
\end{figure}

\paragraph{Evaluation and Benchmarking:} In this final pattern demonstration, we
show how benchmarking a RAG system can be achieved.
Figure~\ref{lst:bench-ragsystem} depicts an intuitive benchmarking pattern
where an \class{Benchmarker}{evals} runs the desired \class{Benchmark}{evals}
using the chosen \class{EvaluationMetric}{evals}.

\begin{figure}[ht!]
\begin{mdframed}[linewidth=1pt]
\begin{Verbatim}[fontsize=\small]
... # Keep code from Figures 1 and 2
import fed_rag.evals.benchmarks as benchmarks
from fed_rag.evals import (
  Benchmarker,
  ExactMatchEvaluationMetric,
)

benchmarker = Benchmarker(rag_system=rag_system)
mmlu = benchmarks.HuggingFaceMMLU(streaming=True)
metric = ExactMatchEvaluationMetric()

# Run benchmark with first 3 examples only
result = benchmarker.run(
  benchmark=mmlu,
  metric=metric,
  is_streaming=True,
  num_examples=3, 
  agg="avg",
)
\end{Verbatim}
\end{mdframed}
\vspace{-3ex}
\caption{Benchmarking with FedRAG.}
\label{lst:bench-ragsystem}
\end{figure}

These patterns demonstrate the consistent API design of FedRAG, enabling users
to seamlessly transition between RAG system development, central and decentralized fine-tuning, and
evaluation with minimal code changes.

\subsection{Integrations}

In this section, we briefly outline the existing integrations to popular tools
and frameworks within the RAG ecosystem. Of the integrations listed in
Table~\ref{tab:integrations}, only the LlamaIndex integration had not been
represented in the preceding patterns of
Figures~\ref{lst:ragsystem}--\ref{lst:bench-ragsystem}. FedRAG supports a bridge
to convert a \lstinline|RAGSystem| object to a LlamaIndex RAG system equivalent,
thus enabling users to leverage their powerful inference features and ecosystem.
These integrations allow FedRAG users to leverage existing tools while gaining
RAG fine-tuning capabilities, aligning with the \emph{Work With Your Tools}
design principle in Section~\ref{sec:design-principles}.

\begin{table}[ht]
\centering
  \footnotesize
  \begin{tabular}{@{}ll@{}}
  \toprule
  Library & Integration \\
  \midrule
    \textbf{\lstinline{HuggingFace}} & Generators, retrievers, datasets \\
    \textbf{\lstinline{Unsloth}} & Fast fine-tuning of generators \\
    \textbf{\lstinline{Qdrant}} & Storage solution for knowledge \\
    \textbf{\lstinline{LlamaIndex}} & Bridge to inference object \\
  \bottomrule
  \end{tabular}
  \caption{Currently supported integrations in FedRAG.}
  \label{tab:integrations} 
\end{table}

\section{Future Work and Conclusions}

In this paper, we introduced FedRAG, a framework for fine-tuning RAG systems
across both centralized and federated architectures that offers state-of-the-art
fine-tuning methods and fills a critical gap within the RAG
ecosystem. In Appendix~\ref{app:rag-specs}, a lightweight experiment is presented. The results confirm that the framework can be used to successfully and
flexibly execute RAG fine-tuning tasks. The experimental code and a containerized image of the knowledge store is released with this paper to facilitate reproducibility.

In terms of future development, we have several exciting and impactful additions on the development roadmap. For example, an MCP RAG
system and companion MCP knowledge store will soon be integrated into the
framework, which will pave the way for studying the effects of adapting RAG
systems to knowledge provided by third-party MCP providers. Additional
high-priority development items are presented in Table~\ref{tab:roadmap} of
Appendix~\ref{app:roadmap}. We are eager to continue the development of FedRAG and believe that it will enable
researchers and practitioners to more easily explore advanced RAG fine-tuning
techniques in both centralized and federated settings.

\bibliography{references}
\bibliographystyle{plainnat}

\newpage
\appendix
\onecolumn
\section{Example: RA-DIT}\label{app:rag-specs}

In their work,~\citet{lin2023ra} conducted various experiments studying the
effectiveness of RALT and LSR fine-tuning methods. Their experiments revealed
that applying RALT or LSR individually leads to performance gains, but the
greatest gain comes after applying both RALT and LSR in succession. They termed
this combination of fine-tuning techniques Retrieval-Augmented Dual Instruction
Tuning (RA-DIT). In order to illustrate the potential of the FedRAG framework, we aim to reproduce a lightweight version of their experiments. The rest of this
appendix outlines the RAG system specifications, details on fine-tuning as well as
evaluation setup, and finally the results of the experiments.

\subsection{RAG System}

\subsubsection{Knowledge Store \& Retriever}

We use text chunks from the December 2021 Wikipedia dump released by~\citet{izacard2022few}. This release includes two files, \texttt{infobox.jsonl}
and \texttt{text-list-100-sec.jsonl}, which can be downloaded from the
\texttt{facebookresearch/atlas} GitHub
repository.\footnote{\url{https://github.com/facebookresearch/atlas}} For the
knowledge store, we use the first 10M text passages provided in
\texttt{text-list-100-sec.jsonl} with no further preprocessing with the
exception of concatenating the \texttt{title}, \texttt{section}, and
\texttt{text} fields for each passage.

For the retriever, we use DRAGON+~\cite{lin2023train}. More specifically, we use
the \texttt{SentenceTransformer} versions of this dual-encoder model available
on
HuggingFace.\footnote{\url{https://huggingface.co/nthakur/dragon-plus-query-encoder}}$^{,}$\footnote{\url{https://huggingface.co/nthakur/dragon-plus-context-encoder}}
The context encoder of DRAGON+ is used to encode the 10M text chunks prior to
loading the embeddings into the knowledge store.

\subsubsection{Generator}

For the generator LLM, we use a quantized (4-bit)
Llama2-7B~\cite{touvron2023llama}. We specifically use the official version of
this model available on
HuggingFace,\footnote{\url{https://huggingface.co/meta-llama/Llama-2-7b-hf}} with the
\texttt{load\_in\_4bit} parameter set to \texttt{True}.

\subsection{Fine-tuning \& Evaluation}

For the fine-tuning dataset, we use Web Questions~\cite{berant2013semantic}
available on
HuggingFace.\footnote{\url{https://huggingface.co/datasets/stanfordnlp/web\_questions}}
We apply QLoRA~\cite{dettmers2023qlora} fine-tuning with only the RALT objective
by making use of a \class{HuggingFaceTrainerForRALT}{trainers} object
and supplying it a \class{HFPeftModelGenerator}{generators}. The latter is used
to load a \texttt{PeftModel} available on
HuggingFace.\footnote{\url{https://huggingface.co/Styxxxx/llama2\_7b\_lora-quac}}

For evaluation, we use the \texttt{test} split of the \texttt{global\_facts} subset of
MMLU~\cite{hendrycks2020measuring} available on
HuggingFace, which has exactly $100$ data points.\footnote{\url{https://huggingface.co/datasets/cais/mmlu}} Similar to
\citet{lin2023ra}, we apply 5-shot in-context learning for each evaluation example. The few-shot examples are randomly drawn from the \texttt{validation} split and held fixed throughout evaluation. For performance measurement,
we use the exact match metric. 

\subsection{Results}

We report the results of two separate fine-tuning runs in
Table~\ref{tab:results}.  For comparison, we also report the performance of the
same RAG system but without any fine-tuning applied at all.

\begin{table}[ht]
  \centering
  \footnotesize
  \begin{tabular}{@{}lccc@{}}
  \toprule
  Method & Run 1 & Run 2 & Average \\
  \midrule
  Without fine-tuning & $17.0$ & $27.0$ & $22.0$ \\
  RALT fine-tuning & $\mathbf{27.0}$ & $\mathbf{34.0}$ & $\mathbf{30.5}$ \\
  \bottomrule
  \end{tabular}
  \caption{RA-DIT inspired experiment demonstrating the effect of RALT
  fine-tuning on exact match performance for the MMLU (\texttt{global\_facts})
  benchmark.}\label{tab:results}
\end{table}

The results of our lightweight experiment corroborate the findings
of~\citet{lin2023ra} and align with results from other
work~\cite{zhang2024raft,chen2025research}. That is, RAG fine-tuning can lead to
significant performance gains. Note that the observed variability in runs is due
to the sampling parameters used for generation.

\section{Development Roadmap}\label{app:roadmap}

In this section, we provide a portion of our development roadmap that includes
the high-priority items, which we deem to be highly impactful for our users.

\begin{table}[ht]
  \centering
  \begin{tabular}{@{}p{0.3\linewidth}p{0.65\linewidth}@{}}
  \toprule
  Item & Description \\
  \midrule
  \textbf{MCP knowledge store} & An MCP client that can be used to receive
  knowledge context from third-party MCP servers. \\
  
  \textbf{MCP RAG system} & A specialized RAG system class that is able to retrieve
  knowledge from the MCP Knowledge Store and subsequently supply it to the
  generator. \\

  \textbf{ReSearch generator trainer} & An implementation of
  ReSearch~\cite{chen2025research}, i.e., equipping generator LLMs with reasoning
  infused with search. \\
  
  \textbf{Improved retriever trainers} & New training objectives for language
  model supervised retriever fine-tuning.\\
  
  \textbf{Advanced federated learning} & Support for more advanced federated
  learning techniques. \\
  
  \textbf{LangChain integration} & Bridge to LangChain
  inference objects. \\ 

  \textbf{General optimizations} & Optimizations for batch RAG system querying
  and concurrent benchmarking. \\

  \bottomrule 
  \end{tabular}
\caption{Development roadmap: high-priority items for FedRAG.}\label{tab:roadmap}
\end{table}


\end{document}